\pdfoutput=1

\documentclass[11pt]{article}

\usepackage{naacl2021}

\usepackage{times}
\usepackage{latexsym}
\usepackage{array}
\usepackage{multirow}

\usepackage[T1]{fontenc}

\usepackage[utf8]{inputenc}
\usepackage{textcomp}

\usepackage{microtype}

\usepackage{times}
\usepackage{latexsym}

\usepackage{graphicx}

\usepackage{subcaption}
\usepackage{url}
\usepackage{amsmath, amsthm, amssymb}

\usepackage{paralist}

\usepackage{enumitem}
\usepackage{xspace}

\usepackage{booktabs} 

\usepackage{wasysym} 
\usepackage{array}

\usepackage{tikz}
\usepackage{float}
\usepackage{arydshln}
\usepackage{centernot}
\usepackage{bbm}

\newcommand{\skippedDetails}[1]{}

\usepackage{xcolor}

\definecolor{orange}{rgb}{1,0.5,0}
\definecolor{mdgreen}{rgb}{0.05,0.6,0.05}
\definecolor{mdblue}{rgb}{0,0,0.7}
\definecolor{dkblue}{rgb}{0,0,0.5}
\definecolor{dkgray}{rgb}{0.3,0.3,0.3}
\definecolor{slate}{rgb}{0.25,0.25,0.4}
\definecolor{gray}{rgb}{0.5,0.5,0.5}
\definecolor{ltgray}{rgb}{0.7,0.7,0.7}
\definecolor{purple}{rgb}{0.7,0,1.0}
\definecolor{lavender}{rgb}{0.65,0.55,1.0}

\newcommand{\alphaOne}{$\alpha_1$\xspace}
\newcommand{\alphaTwo}{$\alpha_2$\xspace}
\newcommand{\alphaThree}{$\alpha_3$\xspace}

\newcommand{\rowKey}[1]{\textit{#1}\xspace}
\newcommand{\entityType}[1]{\textcolor{mdgreen}{\textsc{#1}}\xspace}

\newcommand{\exampleParagraph}[2]{\framebox{\parbox{0.95\linewidth}{\paragraph{#1} 
{\footnotesize #2}}}}

\title{Incorporating External Knowledge to Enhance Tabular Reasoning}

\author{J. Neeraja\thanks{*The first two authors contributed equally to the work. The first author was a remote intern at University of Utah during the work.} \\
   IIT Guwahati\\
  \texttt{jneeraja@iitg.ac.in} \\\And
  Vivek Gupta\footnotemark[1] \\
  University of Utah \\
  \texttt{vgupta@cs.utah.edu} \\ \And
  Vivek Srikumar \\
  University of Utah \\
  \texttt{svivek@cs.utah.edu}}

\begin{document}
\maketitle

\begin{abstract}
Reasoning about tabular information presents unique challenges to modern NLP approaches which largely rely on pre-trained contextualized embeddings of text. In this paper, we study these challenges through the problem of tabular natural language inference. We propose easy and effective modifications to how information is presented to a model for this task. We show via systematic experiments that these strategies substantially improve tabular inference performance.
\end{abstract}

\section{Introduction}
\label{sec:introduction&motivation}

Natural Language Inference (NLI) is the task of determining if a hypothesis sentence can be inferred
as true, false, or undetermined given a premise sentence~\cite{dagan2013recognizing}. Contextual sentence embeddings such as BERT~\cite{devlin2019bert} and RoBERTa~\cite{liu2019roberta}, applied to large datasets such as  SNLI~\cite{snli:emnlp2015} and MultiNLI~\cite{N18-1101}, have led to near-human performance of NLI systems.

In this paper, we study the harder problem of reasoning about \emph{tabular} premises, as instantiated in datasets such as TabFact~\cite{chen2019tabfact} and InfoTabS~\cite{gupta-etal-2020-infotabs}. This problem is similar to standard NLI, but the premises are Wikipedia tables rather than sentences. Models similar to the best ones for the standard NLI datasets struggle with tabular inference. 
Using the InfoTabS dataset as an example, we present a focused study that investigates
\begin{inparaenum}[(a)]
\item the poor performance of existing models, 
\item connections to information deficiency in the tabular premises, and,
\item simple yet effective mitigatations for these problems.
\end{inparaenum}

\begin{figure}[t]
  \centering
  {
  \footnotesize
  \begin{center}
    \begin{tabular}{>{\raggedright}p{0.25\linewidth}p{0.6\linewidth}}
      \toprule
      \multicolumn{2}{c}{\bf New York Stock Exchange}                                                    \\
      \midrule
      {\bf Type} & Stock exchange \\
                                              
      {\bf Location}                & New York City, New York, U.S.                                                  \\ 
      {\bf Founded}           & May 17, 1792; 226 years ago          \\  
      {\bf Currency}                  & United States dollar                             \\ 
      {\bf No. of listings}      & 2,400                                            \\  
     {\bf Volume}             & US\$20.161 trillion (2011)                                                 \\ 
      \bottomrule
    \end{tabular}
  \end{center}
}
  {\footnotesize
    \begin{enumerate}[nosep]
    \item[H1:] NYSE has fewer than 3,000 stocks listed.
    \item[H2:] Over 2,500 stocks are listed in the NYSE.
    \item[H3:] S\&P 500 stock trading volume is over \$10 trillion.
    \end{enumerate}}
    \caption{A tabular premise example. The hypotheses H1 is entailed by it, H2 is a contradiction and H3 is neutral i.e. neither entailed nor contradictory.}
  \label{fig:example}
\end{figure}

We use the table and hypotheses in Figure~\ref{fig:example} as a running example through this paper, and refer to the left column as its keys.\footnote{Keys in the InfoTabS tables are similar to column headers in the TabFact database-style tables.}
Tabular inference is challenging for several reasons:
\begin{inparaenum}[(a)]
\item \textbf{Poor table representation}: The table does not explicitly state the relationship between the keys and values.
\item \textbf{Missing implicit lexical knowledge} due to limited training data: This affects interpreting words like \textit{`fewer'}, and \textit{`over'} in H1 and H2 respectively.
\item \textbf{Presence of distracting information}: All keys except \rowKey{No. of listings} are  unrelated to the hypotheses H1 and H2.
\item \textbf{Missing domain knowledge about keys}: We need to interpret the key \rowKey{Volume} in the financial context for this table.
\end{inparaenum}

In the absence of large labeled corpora, any modeling strategy needs to explicitly address these problems. In this paper, we propose effective approaches for addressing them, and show that they lead to substantial improvements in prediction quality, especially on adversarial test sets. This focused study makes the following  contributions:

\begin{enumerate}[nosep]
    \item We analyse why the existing state-of-the-art BERT class models struggle on the challenging task of NLI over tabular data.
    \item We propose solutions to overcome these challenges via simple modifications to inputs using existing language resources.
    \item Through extensive experiments, we show significant improvements to model performance,  especially on challenging adversarial test sets. 
\end{enumerate}

The updated dataset, along with associated scripts, are available at \url{https://github.com/utahnlp/knowledge_infotabs}.


\section{Challenges and Proposed Solutions}
\label{sec:problem&solution}

We examine the issues highlighted in \S\ref{sec:introduction&motivation} and propose simple solutions to mitigate them below.

\paragraph{Better Paragraph Representation (BPR):}

One way to represent the premise table is to use a universal template to convert each row of the table into sentence which serves as input to a BERT-style model. \citet{gupta-etal-2020-infotabs} suggest that in a table titled \texttt{t}, a row with key \texttt{k} and value \texttt{v} should be converted to a sentence using the template: ``The \texttt{k} of \texttt{t} are \texttt{v}.'' Despite the advantage of simplicity, the approach produces ungrammatical sentences. In our example, the template converts the \rowKey{Founded} row to the sentence \textit{``The Founded of New York Stock Exchange are May 17, 1792; 226 years ago.''}.

We note that keys are associated with values of specific entity types such as
\entityType{Money}, \entityType{Date}, \entityType{Cardinal}, and \entityType{Bool}, and the entire table itself has a category. Therefore, we propose type-specific templates, instead of using the universal one.\footnote{The construction of the template sentences based on entity type is a one-time manual step.}  In our example, the table category is \rowKey{Organization} and the key \rowKey{Founded} has the type \entityType{Date}. A better template for this key is ``\texttt{t} was \texttt{k} on \texttt{v}'', which produces the more grammatical sentence \textit{"New York Stock Exchange was Founded on May 17, 1792; 226 years ago."}. Furthermore, we observe that including the table category information i.e. \textit{``New York Stock Exchange is an Organization.''} helps in better premise context understanding.\footnote{This category information is provided in the InfoTabS and TabFact datasets. For other datasets, it can be inferred easily by clustering over the keys of the training tables.} Appendix~\ref{sec:templateexamples} provides more such templates.  

\paragraph{Implicit Knowledge Addition (KG implicit):}

Tables represent information \emph{implicitly}; they do not employ connectives to link their cells. As a result, a model trained only on tables struggles to make lexical inferences about the hypothesis, such as 
the difference between the meanings of \textit{`before'} and \textit{`after'},
and the
function of negations. 
This is surprising, because the models have the benefit of being pre-trained on large textual corpora.

Recently, \citet{andreas-2020-good} and \citet{pruksachatkun2020intermediate} showed that we can pre-train models on specific tasks  to incorporate such implicit knowledge. \citet{eisenschlos2020understanding} use pre-training on synthetic data to improve the performance on the TabFact dataset. Inspired by these, we first train our model on the large, diverse and \emph{human-written} MultiNLI dataset. 
Then, we fine tune it to the InfoTabS task. Pre-training with MultiNLI data exposes the model to diverse lexical constructions. Furthermore, it increases the training data size by $433$K (MultiNLI) example pairs. This makes the representation better tuned to the NLI task, thereby leading to better generalization. 

\paragraph{Distracting Rows Removal (DRR)}
Not all premise table rows are necessary to reason about a given hypothesis. In our example, for the hypotheses H1 and H2, the row corresponding to the key \rowKey{No. of listings} is sufficient to decide the label for the hypothesis. The other rows are an irrelevant distraction.
Further, as a practical concern, when longer tables are encoded into sentences
as described above, the resulting number of tokens is more than the input size
restrictions of existing models, leading to useful rows potentially being cropped. 
Appendix~\ref{sec:appendixexample} shows one such example on the InfoTabS. Therefore, it becomes important to prune irrelevant rows.

To identify relevant rows, we employ a simplified version of the alignment algorithm used by \citet{yadav2019alignment,yadav-etal-2020-unsupervised} for retrieval in reading comprehension.

First, every word in the hypothesis sentence is aligned with the most similar word in the table sentences using cosine similarity. We use fastText~\cite{joulin2016fasttext,mikolov2018advances} embeddings for this purpose, which preliminary experiments revealed to be better than other embeddings. Then, we rank rows by their similarity to the hypothesis, by aggregating similarity over content words in the hypothesis. \citet{yadav2019alignment} used inverse document frequency for weighting words, but we found that simple stop word pruning was sufficient. We took the top $k$ rows by similarity as the pruned representative of the table for this hypothesis.
The hyper-parameter $k$ is selected by tuning on a development set. Appendix~\ref{sec:appendixfastText} gives more details about these design choices. 

\paragraph{Explicit Knowledge Addition (KG explicit):}

We found that adding \emph{explicit} information to enrich keys improves a
model's ability to disambiguate and understand them. We expand the pruned table
premises with contextually relevant key information from existing resources such
as WordNet (definitions) or Wikipedia (first sentence, usually a
definition).\footnote{Usually multi-word keys are absent in WordNet, in this
  case we use Wikipedia. The WordNet definition of each word in the key is used
  if the multi-word key is absent in Wikipedia.}


To find the best expansion of a key, we use the sentential form of a row to
obtain the BERT embedding (on-the-fly) for its key. We also obtain the BERT
embeddings of the same key from WordNet examples (or Wikipedia
sentences).\footnote{We prefer using WordNet examples over definition for BERT
  embedding because (a) an example captures the context in which key is used,
  and (b) the definition may not always contain the key tokens.} Finally, we
concatenate the WordNet definition (or the Wikipedia sentence) corresponding to the highest key embedding similarity to the table. As we want the contextually relevant definition of the key, we use the BERT embeddings rather than non-contextual ones (e.g., fastText). For example, the key \rowKey{volume} can have different meanings in various contexts. For our example, the contextually best definition is \textit{``In capital markets, \textbf{volume}, is the total number of a security that was traded during a given period of time.''} rather than the other definition \textit{``In thermodynamics, the \textbf{volume} of a system is an extensive parameter for describing its thermodynamic state.''}.


\section{Experiment and Analysis}
\label{sec:experiment&analysis} 

Our experiments are designed to study the research question: \emph{Can today's large pre-trained models exploit the information sources described in \S\ref{sec:problem&solution} to better reason about tabular information?} 

\subsection{Experimental setup} 

\paragraph{Datasets} Our experiments uses InfoTabS, a tabular inference dataset from \citet{gupta-etal-2020-infotabs}. The dataset is heterogeneous in the types of tables and keys, and relies on background knowledge and common sense. Unlike the TabFact dataset~\cite{chen2019tabfact}, it has all three inference labels, namely entailment, contradiction and neutral. Importantly, for the purpose of our evaluation, it has three test sets. In addition to the usual development set and the test set (called \alphaOne), the dataset has two adversarial test sets: a contrast set \alphaTwo that is lexically similar to \alphaOne, but with minimal changes in the hypotheses and flip entail-contradict label, and a zero-shot set \alphaThree which has long tables from different domains with little key overlap with the training set. 

\paragraph{Models} For a fair comparison with earlier baselines, we use RoBERTa-large (RoBERTa$_{L}$) for all our experiments. We represent the premise table by converting each table row into a sentence, and then appending them into a paragraph, i.e. the \textit{Para} representation of \citet{gupta-etal-2020-infotabs}. 

\paragraph{Hyperparameters Settings\footnote{Appendix~\ref{sec:hyperaparameters} has more details about hyperparameters.}} For the distracting row removal (+DRR) step, we have a hyper-parameter $k$. We experimented with $k \in \{2,3,4,5,6\}$, by predicting on +DRR development premise on model trained on orignal training set (i.e. BPR), as shown in Table \ref{tab:hyperparameterk}. The development accuracy increases significantly as $k$ increases from $2$ to $4$ and then from  $4$ to $6$, increases marginally ($~1.5\%$ improvement). Since our goal is to remove distracting rows, we use the lowest hyperparameter with good performance i.e. $k=4$.\footnote{Indeed, the original InfoTabs work points out that no more than four rows in a table are needed for any hypothesis.}.

\begin{table}[h]
\centering
\renewcommand{\tabcolsep}{2.6pt}
\small
\begin{tabular}{ccccccc}
\toprule
Train & Dev & k = 2 & k = 3 & k = 4 & k = 5 & k = 6\\
\midrule
BPR & DRR & 71.72 & 74.83 & 77.50 & 78.50& 79.00\\
\bottomrule
\end{tabular}
\caption{Dev accuracy on increasing hyperparameter $k$.}
\label{tab:hyperparameterk}
\end{table}

\subsection{Results and Analysis}
Table~\ref{tab:pararesults} shows the results of our experiments.

\begin{table}[h]
\small
\centering
\begin{tabular}{lrrrr}
\toprule
Premise & Dev & \alphaOne & \alphaTwo & \alphaThree \\
\midrule
Human &\bf 79.78 & \bf 84.04 &\bf 83.88 &\bf 79.33 \\
Para &	75.55 & 74.88	& 65.55 & 64.94 \\ 
\midrule 
BPR	& 76.42	& 75.29 & 66.50 & 64.26 \\
+KG implicit	&\bf 79.57	&78.27 & 71.87 & 66.77 \\
+DRR & 78.77	& 78.13 & 70.90 & 68.98 \\
+KG explicit	&79.44	&\bf 78.42 & \bf 71.97 & \bf 70.03 \\
\bottomrule
\end{tabular}
\caption{Accuracy with the proposed modifications on the Dev and test
  sets. Here, + represents the change with respect to the previous row. Reported
  numbers are the average over three random seed runs with standard deviation of $0.33$ (+KG explicit), $0.46$ (+DRR), $0.61$ (+KG implicit), $0.86$ (BPR), over all sets. All improvements are statistically significant with $p<0.05$, except \alphaOne for BPR representation w.r.t to Para (Original). Here the Human and Para results are taken from~\citet{gupta-etal-2020-infotabs}.}
\label{tab:pararesults} 
\end{table}

\paragraph{BPR} 
As shown in Table \ref{tab:pararesults}, with BPR, we observe that the RoBERTa$_{L}$ model improves performance on all dev and test sets except \alphaThree. There are two main reasons behind this poor performance on \alphaThree.

First, the zero-shot \alphaThree data includes  unseen  keys. The number of keys common to \alphaThree and the training set is $94$, whereas for, dev, \alphaOne and \alphaTwo it is $334$, $312$, and $273$ respectively (i.e., 3-5 times more). 
Second, despite being represented by better sentences, due to the input size restriction of RoBERTa$_L$ some relevant rows are still ignored.

\paragraph{KG implicit}
We observe that \emph{implicit} knowledge addition via MNLI pre-training helps the model reason and generalize better. From Table \ref{tab:pararesults}, we can see significant performance improvement in the dev and all three test sets. 

\paragraph{DRR} This leads to significant improvement in the \alphaThree set. We attribute this to two primary reasons: 
First, \alphaThree tables are longer ($13.1$ keys per table on average,
vs. $~8.8$ keys on average in the others), and DRR is important to avoid
automatically removing keys from the bottom of a table due to the limitations in RoBERTa$_L$ model's input size. Without these relevant rows, the model incorrectly predicts the neutral label.
Second, \alphaThree is a zero-shot dataset and has significant proportion of unseen keys which could end up being noise for the model. The slight decrease in performance on the dev, \alphaOne and \alphaTwo sets can be attributed to model utilising spurious patterns over irrelevant keys for prediction.\footnote{Performance drop of dev and \alphaTwo is also marginal i.e. (dev: 79.57 to 78.77, \alphaOne: 78.27 to 78.13, \alphaTwo: 71.87 to 70.90), as compared to InfoTabS WMD-top3 i.e (dev: 75.5 to 72.55,\alphaOne: 74.88 to 70.38, \alphaTwo: 65.44 to  62.55), here WMD-top3 performance numbers are taken from ~\citet{gupta-etal-2020-infotabs}.} We validated this experimentally by testing the original premise trained model on the DRR test tables. Table~\ref{tab:testonpruneappendix} in the Appendix~\ref{sec:hyperaparameters} shows that without pruning, the model focuses on irrelevant rows for prediction.

\paragraph{KG explicit} 
With \emph{explicit} contextualized knowledge about the table keys, we observe a marginal improvement in dev, \alphaOne test sets and a significant performance gain on the \alphaTwo and \alphaThree test sets. Improvement in the \alphaThree set shows that adding external knowledge helps in the zero-shot setting. With \alphaTwo, the model can not utilize spurious lexical correlations\footnote{The hypothesis-only baseline for \alphaTwo is 48.5$\%$ vs. \alphaOne: 60.5 $\%$ and dev: 60.5 $\%$ \cite{gupta-etal-2020-infotabs}} due to its adversarial nature, and is forced to use the relevant keys in the premise tables, thus adding explicit information about the key improves performance more for \alphaTwo than \alphaOne or dev. Appendix \ref{sec:appendixexample} shows some qualitative examples. 

\subsection{Ablation Study}

We  perform an ablation study as shown in table~\ref{tab:ablall}, where instead of doing all modification sequentially one after another (+), we do only one modification at a time to analyze its effects. 

Through our ablation study we observe that: \begin{inparaenum}[(a)] \item \textbf{DRR} improves performance on the dev, \alphaOne, and \alphaTwo sets, but slightly degrades it on the \alphaThree set. The drop in performance on \alphaThree is due to spurious artifact deletion as explained in details in Appendix~ \ref{sec:appendixartifact}.
\item \textbf{KG explicit} gives performance improvement in all sets. Furthermore, there is significant boost in performance of the adversarial \alphaTwo and \alphaThree sets.\footnote{The KG explicit step is performed only for relevant  keys (after DRR).}
\item Similarly, \textbf{KG implicit} shows significant improvement in all test sets. The large improvements on the adversarial sets \alphaTwo and \alphaThree sets, suggest that the model can now reason better. Although, implicit knowledge provides most performance gain, all modifications are needed to obtain the best performance for all sets (especially on the \alphaThree set).\footnote{We show in Appendix~\ref{sec:strucexepriments}, Table~\ref{tab:strucresults}, that implicit knowledge addition to a non-sentential table representation i.e. Struc \cite{chen2019tabfact,gupta-etal-2020-infotabs} leads to performance improvement as well.}

\begin{table}[!htbp]
\small
\centering
\begin{tabular}{lcccc}
\toprule
Premise & Dev & \alphaOne & \alphaTwo & \alphaThree \\
\midrule
Para &	75.55 & 74.88	& 65.55 & 64.94 \\ 
\midrule 
DRR & 76.39	& 75.78 & 67.22 & 64.88 \\
KG explicit & 77.16 & 75.38 & 67.88 & 65.50 \\
KG implicit &\bf 79.06	&\bf 78.44 &\bf 71.66 &\bf 67.55 \\
\bottomrule
\end{tabular}
\caption{Ablation results with individual modifications.}

\label{tab:ablall}
\end{table}

\end{inparaenum}


\section{Comparison with Related Work}
\label{sec:related_work}

Recently, there have been many papers which study several NLP tasks on semi-structured tabular data. These include tabular NLI and fact verification tasks such as TabFact~\cite{chen2019tabfact}, and InfoTabS~\cite{gupta-etal-2020-infotabs}, 
various question answering and semantic parsing tasks~\cite[][\emph{inter alia}]{pasupat2015compositional,krishnamurthy2017neural,Abbas2016WikiQAA,sun2016table,chen2020hybridqa,lin2020bridging}, and
table-to-text generation and its evaluation~\cite[e.g.,][]{parikh2020totto,radev2020dart}.
%
Several, models for better representation of tables such as
TAPAS~\cite{herzig-etal-2020-tapas}, TaBERT \cite{yin-etal-2020-tabert}, and
TabStruc~\cite{zhang-etal-2020-table} were recently proposed.
\citet{yu2018spider,yu2020grappa} and \citet{eisenschlos2020understanding} study pre-training for improving tabular inference, similar to our MutliNLI pre-training. 

The proposed modifications in this work are simple and intuitive. Yet, existing table reasoning papers have not studied the impact of such input modifications. Furthermore, much of the recent work focuses on building sophisticated neural models, without explicit focus on how these models (designed for raw text) adapt to the tabular data. In this work, we argue that instead of relying on the neural network to \textit{``magically''} work for tabular structures, we should carefully think about the representation of semi-structured data, and the incorporation of both implicit and explicit knowledge into neural models. Our work highlights that simple pre-processing steps are important, especially for better generalization, as evident from the significant improvement in performance on adversarial test sets with the same RoBERTa models. We recommend that these pre-processing steps should be standardized across table reasoning tasks. 



\section{Conclusion \& Future Work}
\label{sec:conclusion}

We introduced simple and effective modifications that rely on introducing additional knowledge to improve tabular NLI. These modifications governs what information is provided to a tabular NLI and how the given information is presented to the model. We presented a case study with the recently published InfoTabS dataset and showed that our proposed changes lead to significant improvements. Furthermore, we also carefully studied the effect of these modifications on the multiple test-sets, and why a certain modification seems to help a particular adversarial set.

We believe that our study and proposed solutions will be valuable to researchers
working on question answering and generation problems involving both tabular and textual inputs, such
as tabular/hybrid question answering and table-to-text generation, especially
with difficult or adversarial evaluation.  Looking ahead, our work can be extended to include explicit knowledge for hypothesis tokens as well. To increase robustness, we can also integrate structural constraints via data augmentation through NLI training. Moreover, we expect that structural information such as position encoding could also help better represent tables.


\section*{Acknowledgements}
\label{sec:aknowledgement}

We thank members of the Utah NLP group for their valuable insights and
suggestions at various stages of the project; and reviewers their helpful
comments. We also thank the support of NSF grants \#1801446 (SATC) and \#1822877
(Cyberlearning) and a generous gift from Verisk Inc.



\bibliography{custom}
\bibliographystyle{acl_natbib}

\clearpage
\appendix

\section{BPR Templates}
Here, we are listing down some of the diverse example templates we have framed.
\label{sec:templateexamples}
\begin{itemize}
\item {For the table category \rowKey{Bus/Train Lines} and key \rowKey{Disabled access} with \entityType{Bool} value YES, follow template: "\textit{t} has \textit{k}."}

\noindent \exampleParagraph{Orignal Premise Sentence}{\textit{``The Disabled access of Tukwila International Boulevard Station are Yes.''}}
\noindent \exampleParagraph{BPR Sentence}{\textit{``Tukwila International Boulevard Station has Disabled access.''}}

\item {For the table category \rowKey{Movie} and key \rowKey{Box office} with \entityType{Money} type, follow template: "In the \textit{k}, \textit{t} made \textit{v}."}

\noindent \exampleParagraph{Orignal Premise Sentence}{\textit{``The Box office of Brokeback Mountain are \$178.1 million.''}}

\noindent \exampleParagraph{BPR Sentence}{\textit{``In the Box office, Brokeback Mountain made \$178.1 million.''}}
\item {For the table category \rowKey{City} and key \rowKey{Total} with \entityType{Cardinal} type, follow template: "The \textit{k} area of \textit{t} is \textit{v}."}

\noindent \exampleParagraph{Orignal Premise Sentence}{\textit{``The Total of Cusco are 435,114.''}}
\noindent \exampleParagraph{BPR Sentence}{\textit{``The Total area of Cusco is 435,114.''}}

\item {For the table category \rowKey{Painting} and key \rowKey{Also known as}, follow template: "The \textit{k} area of \textit{t} is \textit{v}."}

\noindent \exampleParagraph{Orignal Premise Sentence}{\textit{``The Also known as of Et in Arcadia ego are Les Bergers d'Arcadie.''}}
\noindent \exampleParagraph{BPR Sentence}{\textit{``Et in Arcadia ego is Also known as Les Bergers d'Arcadie.''}}

\item {For the table category \rowKey{Person} and key \rowKey{Died} with \entityType{Date} type , follow template: "\textit{t} \textit{k} on \textit{v}."}

\noindent \exampleParagraph{Orignal Premise Sentence}{\textit{``The Died of Jesse Ramsden are November 1800 (1800-11-05)  (aged 65)  Brighton, Sussex.''}}
\noindent \exampleParagraph{BPR Sentence}{\textit{``Jesse Ramsden Died on 5 November 1800 (1800-11-05)  (aged 65)  Brighton, Sussex.''}}

\end{itemize}

\section{DRR: fastText and Binary weighting}
\label{sec:appendixfastText}
\paragraph{fastText:} For word representation, \cite{yadav2019alignment} have used BERT and Glove embeddings. In our case, we prefer to use  fastText word embeddings over Glove because fastText embedding uses sub-word information which helps in capturing different variations of the context words. Furthermore, fastText embeddings is also as better choice than BERT for our task because 
\begin{inparaenum}
\item Firstly, we are embedding single sentential form of diverse rows instead of longer context similar paragraphs,
\item Secondly, all words (especially keys) of the rows across all the tables are used only in one context, whereas BERT is useful when same word is used with different contexts across paragraphs, 
\item Thirdly, in all tables, the number sentences to select from is bounded by maximum rows in the table, which is a small number (8.8 in train, dev, \alphaOne, \alphaTwo and 13.1 in \alphaThree), and
\item Lastly, using fastText is much faster to compute than BERT for obtaining embeddings.
\end{inparaenum}

\paragraph{Binary weighting:} Since, we are embedding single sentential form of diverse rowsinstead of longer context related paragraphs, we found that using binary weighting $0$ for stop words and $1$ for others is more effective than the idf weighting, which is useful only for longer paragraph context with several lexical terms.

\section{Hyperparameters $k$ vs test-sets accuracy}
\label{sec:hyperaparameters}
We also trained a model both train and tested on the DRR table premise for increasing values of the hyper parameter $k$, as shown in  Table \ref{tab:hyperparameterk}. We also test the model trained on the entire para on pruned para with increasing value of hyperparameters $k$ $\in$ $\{2,3,4,5,6\}$ for the test sets \alphaOne, \alphaTwo, and \alphaThree. In all cases, except \alphaThree, the performance with larger $k$ is better. The increase in performance, even with $k$ $>$ $4$, shows that the model is using more then required keys for prediction. Thus, the model is utlising the spurious pattern in irrelevant rows for the prediction.


\begin{table}[h]
\small
\centering
\renewcommand{\tabcolsep}{2.6pt}
\begin{tabular}{llcccccc}
\toprule
Train & Dev  & k=2 & k=3 & k=4 & k=5 & k=6\\
\midrule
+DRR & +DRR & 77.61 & 77.94 & 78.16 & 78.38 & 79.00\\
BPR & +DRR & 71.72 & 74.83 & 77.50& 78.50& 79.00\\
\bottomrule
\end{tabular}
\caption{Dev accuracy with increasing hyper parameter $k$ trained with both BPR and +DRR table.}
\label{tab:hyperparameterkappendix}
\end{table}

\begin{table}[!htbp]
\small
\centering
\begin{tabular}{lccc}
\toprule
$k$ & \alphaOne & \alphaTwo & \alphaThree \\
\midrule
2 & 71.44 & 67.33 & 64.83\\
3 & 75.05	& 69.33 & 67.33 \\ 
4 & 77.72 & 69.83 & 68.22 \\
5 & 77.77 & 70.28 & \bf69.28 \\
6 & \bf 77.77 &\bf 70.77 & 69.22 \\
\bottomrule
\end{tabular}
\caption{Accuracy of model trained with orignal table but tested with DRR table with increasing hyper parameter $k$ on all test sets.}
\label{tab:testonpruneappendix}
\end{table}

\section{TabFact Representation Experiment}
\label{sec:strucexepriments}
Table \ref{tab:strucresults} implicit knowledge addition effect on non-para  \textit{Struc} representation i.e. a key value linearize representation as ``key \texttt{k} : value \texttt{v}'', rows separated by semicolon ``;" \cite{gupta-etal-2020-infotabs,chen2019tabfact}. Here too the implicit knowledge addition leads to improvement in performance on all the sets.

\begin{table}[!htbp]
\small
\centering
\begin{tabular}{lcccc}
\toprule
Premise & Dev & \alphaOne & \alphaTwo & \alphaThree \\
\midrule
Struc &	77.61 & 75.06	& 69.02 & 64.61 \\ 
+ KG implicit &\bf 79.55	&\bf 78.66 &\bf 72.33 &\bf 70.44 \\
\bottomrule
\end{tabular}
\caption{Accuracy on InfoTabS data for Struc representation of Tables. Here, + represents the change with respect to the previous row.}
\label{tab:strucresults}
\end{table}

\section{Artifacts and Model Predictions}
\label{sec:appendixartifact}

In Table \ref{tab:crosspredictions} we show percentage of example which were corrected after modification and vice versa. Surprisingly, there is a small percentage of examples which are predicted correctly earlier with original premise (Para) but predicted wrongly after all the modifications (Mod), although such examples are much lesser than opposite case. We suspect that earlier model was also relying on spurious pattern (artifacts) for correct prediction on these examples earlier, which are now corrupted after the proposed modifications. Hence, the new model struggle to predict correctly on such examples.

\begin{table}[h]
\small
\centering
\begin{tabular}{llcccc}
\toprule
Para & Mod & Dev & \alphaOne & \alphaTwo & \alphaThree \\
\midrule
\checkmark & $\times$	& 6.77	& 7.83 & 9.27 & 10.01 \\
$\times$ & \checkmark 	&\bf 10.94	&\bf 12.55 &\bf 14.33 & \bf16.05 \\
\bottomrule
\end{tabular}
\caption{Correct vs Incorrect Predictions for Para model ~\cite{gupta-etal-2020-infotabs} and the model after the modifcations (Mod).}
\label{tab:crosspredictions}
\end{table}

In the next section \ref{sec:appendixexample}, we also shows qualitative examples, where modification helps model predict correctly. We also provide some examples via distracting row removal modification, where model fails after modification. 

\section{Qualitative Examples}
\label{sec:appendixexample}

In this section, we provide examples where model is able to predict well after the proposed modifications. We also provide some examples, where model struggles to make the correct prediction after distracting row removal (DRR) modification.

\subsection{BPR}

\exampleParagraph{Original Premise}{ The Birth name of Eva Mendes are Eva de la Caridad Méndez. Eva Mendes was Born on March 5, 1974  (1974-03-05)  (age 44)   Miami, Florida, U.S.. The Occupation of Eva Mendes are Actress, model, businesswoman. The Years active of Eva Mendes are 1998 - present. The Partner(s) of Eva Mendes are Ryan Gosling (2011 - present). The Children of Eva Mendes are 2.}
\noindent \exampleParagraph{\color{black} Better Paragraph Premise} {\textcolor{mdgreen} {Eva Mendes is a person.} The birth name of Eva Mendes is Eva de la Caridad Méndez.  Eva Mendes was born on March 5, 1974 (1974-03-05)  (age 44)   Miami, Florida, U.S..  The occupation of Eva Mendes is Actress, model, businesswoman.  The years active of Eva Mendes was on 1998 - present.  The partner(s) of Eva Mendes is Ryan Gosling (2011 - present). \textcolor{mdgreen}{ The number of children of Eva Mendes are 2.}}
\noindent \exampleParagraph{Hypothesis}{Eva Mendes has two children.}

\paragraph{Result and Explanation}
\begin{table}[!htbp]
\small
    \centering
    \begin{tabular}{c|c} 
    \toprule
     Premise & Label \\ \midrule
     Human Label (Gold) & Entailed  \\
     Orignal Premise & Neutral \\
     +BPR & Entailed \\
    \bottomrule
    \end{tabular}
    \caption{Prediction after BPR. Here, + represents the change with respect to the previous row.}
    \label{tab:bpr}
\end{table}

In this example from \alphaTwo, the model predicts Neutral for this hypothesis with orignal premise. However, forming better sentences by adding the \textit{"number of children are 2"} (highlighted as \textcolor{mdgreen}{ green}) in case of CARDINAL type for the category PERSON helps the model understand the relation and reasoning behind the children and the number two and arrive at the correct prediction of entailment.

\subsection{KG implicit}

\exampleParagraph{Original Premise}{
Janet Leigh is a person. Janet Leigh was born as Jeanette Helen Morrison  (1927-07-06) July 6, 1927  Merced, California, U.S.  Janet Leigh died on October 3, 2004 (2004-10-03) (aged 77)  Los Angeles, California, U.S..  The resting place of Janet Leigh is Westwood Village Memorial Park Cemetery.  The alma mater of Janet Leigh is University of the Pacific.  The occupation of Janet Leigh are Actress, singer, dancer, author.  The years active of Janet Leigh was on 1947-2004.  The political party of Janet Leigh is Democratic.  The spouse(s) of Janet Leigh are John Carlisle (m.  1942; annulled 1942), Stanley Reames (m.  1945;  div.  1949), Tony Curtis (m.  1951;  div.  1962), Robert Brandt (m.  1962).  The children of Janet Leigh are Kelly Curtis, Jamie Lee Curtis.
}

\noindent \exampleParagraph{Hypothesis A}{
Janet Leigh's career spanned \textbf{over} 55 years long.}

\noindent \exampleParagraph{Hypothesis B}{Janet Leigh's career spanned \textbf{under} 55 years long.}

\paragraph{Result and Explanation} 

\begin{table}[!htbp]
\small
    \centering
    \begin{tabular}{c|c} 
    \toprule
     Premise & Label \\ \midrule
     Human Label (Gold) & Entailed  \\
     Orignal Premise & Entailed \\
     + KG implicit & Entailed \\
    \bottomrule
    \end{tabular}
    \caption{Prediction on Hypothesis A. Here, + represents the change with respect to the previous row}
    \label{tab:org}
\end{table}

\begin{table}[!htbp]
\small
    \centering
    \begin{tabular}{c|c} 
    \toprule
     Premise & Label \\ \midrule
     Human Label (Gold) & Contradiction  \\
     Orignal Premise & Entailed \\
     + KG implicit & Contradiction \\
    \bottomrule
    \end{tabular}
    \caption{Prediction on Hypothesis B (from \alphaTwo). Here, + represents the change with respect to the previous row}
    \label{tab:flipped}
\end{table}

In this example from \alphaTwo, the model without implicit knowledge and the model with implicit knowledge addition predict the correct label on the  Hypothesis A. However for Hypothesis B which is an example from \alphaTwo, and originally generated by replacing the word "\textbf{over}" to word "\textbf{under}" in the Hypothesis A and flipping gold label from entail to contradiction, the ealier model which is using artifacts over lexical patterns arrive to predict the original wrong label entail instead of contradiction. On adding implicit knowledge while training, the model is now able to reason rather than relying on artifacts and correctly predicts contradiction. Note, that both hypothesis A and hypothesis B require exactly same reasoning for inference i.e. they are equally hard.

\subsection{KG explicit}

\subsection{DRR}

\exampleParagraph{Original Premise}{ The pronunciation of Fluorine are (FLOOR-een, -in, -yn) and (FLOR-een, -in, -yn).  The allotropes of Fluorine is alpha, beta.  The appearance of Fluorine is gas: very pale yellow , liquid: bright yellow , solid: alpha is opaque, beta is transparent.  The standard atomic weight are, std(f) of Fluorine is 18.998403163(6).  The atomic number (z) of Fluorine is 9. {\color{blue}The group of Fluorine is group 17 (halogens). } The period of Fluorine is period 2.  The block of Fluorine is p-block. The element category of Fluorine is Reactive nonmetal.  The electron configuration of Fluorine is [He] 2s 2  2p 5.  The electrons per shell of Fluorine is 2, 7.  The phase at stp of Fluorine is gas.  The melting point of Fluorine is (F-2) 53.48 K (-219.67 °C, -363.41 °F).  The boiling point of Fluorine is (F 2 ) 85.03 K (-188.11 °C, -306.60 °F).  The density (at stp) of Fluorine is 1.696 g/L.  The when liquid (at b.p.) of Fluorine is 1.505 g/cm 3.  The triple point of Fluorine is 53.48 K, 90 kPa.  The critical point of Fluorine is 144.41 K, 5.1724 MPa.  The heat of vaporization of Fluorine is 6.51 kJ/mol.  The molar heat capacity of Fluorine is C p : 31 J/(mol·K) (at 21.1 °C) , C v : 23 J/(mol·K) (at 21.1 °C).  The oxidation states of Fluorine is -1  (oxidizes oxygen).  The electronegativity of Fluorine is Pauling scale: 3.98. {\color{blue} Fluorine was ionization energies on 1st: 1681 kJ/mol, 2nd: 3374 kJ/mol, 3rd: 6147 kJ/mol, (more).}  The covalent radius of Fluorine is 64 pm.  The van der waals radius of Fluorine is 135 pm.  The natural occurrence of Fluorine is primordial.  The thermal conductivity of Fluorine is 0.02591 W/(m·K).  The magnetic ordering of Fluorine is diamagnetic (-1.2×10 -4 ).  The cas number of Fluorine is 7782-41-4.  The naming of Fluorine is after the mineral fluorite, itself named after Latin  fluo  (to flow, in smelting). \textcolor{mdgreen}{The discovery of Fluorine is André-Marie Ampère (1810).} {\color{blue} The first isolation of Fluorine is Henri Moissan (June 26, 1886). } The named by of Fluorine is Humphry Davy.}

\noindent \exampleParagraph{Distracting Row Removal (DRR)}{ The first isolation of Fluorine is Henri Moissan (June 26, 1886).  The group of Fluorine is group 17 (halogens).  \textcolor{mdgreen}{The discovery of Fluorine is André-Marie Ampère (1810).}  Fluorine was ionization energies on 1st: 1681 kJ/mol, 2nd: 3374 kJ/mol, 3rd: 6147 kJ/mol, (more). }

\noindent \exampleParagraph{Hypothesis} {Flourine was discovered in the 18th century.}

\paragraph{Result and Explanation} 
\begin{table}[!htbp]
\small
    \centering
    \begin{tabular}{c|c} 
    \toprule
       Premise & Label \\ \midrule
         Human Label (Gold) & Contradiction  \\
         Orignal Premise & Neutral \\
         +DRR & Contradiction \\
    \bottomrule
    \end{tabular}
    \caption{Prediction after DRR. Here, + represents the change with respect to the previous row.}
    \label{tab:examplePruning1}
\end{table}

In this example from the \alphaThree set, removing distracting rows (sentence except the one in \textcolor{mdgreen} {green} and {\color{blue} blue}) definitely helps as there are irrelevant distracting noise and also make premise paragraph long beyond BERT maximum tokenization limits. Before DRR is applied, the model predicts neutral due to a) distracting rows and b) required information i.e. relevant keys-rows {highlighted as \textcolor{mdgreen} {green}} being removed due to maximum tokenization limitation (it's second last sentence). However, after DRR, the prune information retained is only the relevant keys {highlighted as \textcolor{mdgreen} {green}} and thus the model is able to predict the correct label. 

\paragraph{Negative Example} In some examples distracting row removal for DRR remove an relevant rows and hence the model failed to predict correctly on the DRR premise, as shown below:\\

\noindent \exampleParagraph{\color{black} Original Premise} {\textcolor{mdgreen} {Et in Arcadia ego is a painting}. Et in Arcadia ego is also known as Les Bergers d'Arcadie.  {\color{blue} The artist of Et in Arcadia ego is Nicolas Poussin.}  The year of Et in Arcadia ego is 1637 - 1638.  {\color{blue}The medium of Et in Arcadia ego is oil on canvas.  The dimensions of Et in Arcadia ego is 87 cm 120 cm (34.25 in 47.24 in).} {\color{red} The location of Et in Arcadia ego is Musee du Louvre. }} 

\noindent \exampleParagraph{\color{black} Distracting Row Removal (DRR)} {{\textcolor{mdgreen} {Et in Arcadia ego is a painting.}}  The artist of Et in Arcadia ego is Nicolas Poussin.  The medium of Et in Arcadia ego is oil on canvas.  The dimensions of Et in Arcadia ego is 87 cm 120 cm (34.25 in 47.24 in). }

\noindent \exampleParagraph{Hypothesis}{ The art piece Et in Arcadia ego is \textbf{stored} in the \textbf{United Kingdom}}.

\paragraph{Result and Explanation} 
\begin{table}[!htbp]
\small
    \centering
    \begin{tabular}{c|c} 
    \toprule
        Premise & Label \\ \midrule
         Human Label (Gold) & Contradiction  \\
         Orignal Premise & Contradiction \\
         +DRR & Neutral \\
    \bottomrule
    \end{tabular}
    \caption{Prediction after DRR. Here, + represents the change with respect to the previous row.}
    \label{tab:examplePruning2}
\end{table}

In this example from the Dev set, the DRR technique used removes the required key "\textit{Location}" (highlighted in {\color{red} red}) from the para representation. Hence, the model here predicts neutral as the information regarding where the painting is stored i.e. "\textit{Location}" is removed in the DRR, which the model require for making the correct inference. While in original para, this information is still present and the model is able to arrive at the correct label. Another interesting observation is RoBERTa$_{L}$ knows Musee du Louvre is a museum in the United Kingdom, showing sign of world-knowledge.

\paragraph{Negative Example} In another negative examples distracting row removal for DRR got the relevant rows correct but still the model failed to predict correct label due to spurious correlation, as shown below:\\

\noindent \exampleParagraph{Original Premise}{Idiocracy is a movie. \textcolor{mdgreen}{ Idiocracy was directed by Mike Judge.}  {\color{blue} Idiocracy was produced by Mike Judge, Elysa Koplovitz, Michael Nelson.}  \textcolor{mdgreen}{ Idiocracy was written by Etan Cohen, Mike Judge.}  Idiocracy was starring Luke Wilson, Maya Rudolph, Dax Shepard.  Idiocracy was music by Theodore Shapiro.  The cinematography of Idiocracy was by Tim Suhrstedt. {\color{blue} Idiocracy was edited by David Rennie.}  The production company of Idiocracy is Ternion.  Idiocracy was distributed by 20th Century Fox.  The release date of Idiocracy is September 1, 2006.  The running time of Idiocracy is 84 minutes.  The country of Idiocracy is United States.  The language of Idiocracy is English.  The budget of Idiocracy is $\$$2-4 million.  In the box office, Idiocracy made $\$$495,303  (worldwide).} 

\noindent \exampleParagraph{Distracting Row Removal (DRR)}{ \textcolor{mdgreen}{Idiocracy was directed by Mike Judge.}  Idiocracy was produced by Mike Judge, Elysa Koplovitz, Michael Nelson.  \textcolor{mdgreen}{Idiocracy was written by Etan Cohen, Mike Judge.}  Idiocracy was edited by David Rennie.} 

\noindent \exampleParagraph{Hypothesis} {Idiocracy was directed and written by the \textbf{same} person.}

\paragraph{Result and Explanation} 
\begin{table}[!htbp]
\small
    \centering
    \begin{tabular}{c|c} 
    \toprule
        Premise & Label \\ \midrule
         Human Label (Gold) & Entailed  \\
         Orignal Premise & Entailed \\
         +DRR & Neutral \\
    \bottomrule
    \end{tabular}
    \caption{Prediction after DRR. Here, + represents the change with respect to the previous row.}
    \label{tab:examplePruning3}
\end{table}

In this example from the Dev set, the model before DRR predicts the correct label but however on DRR, it predicts incorrect label of neutral. Despite the fact that both the relevant rows require for inference (highlighted in \textcolor{mdgreen} {green}) is present after DRR. This shows, that the model is looking at more keys than required in the initial case, which are eliminated in the DRR, which force the model to change it prediction. Thus, model is utilising spurious correlation from irrelevant rows to predict the label.

\noindent  \exampleParagraph{Orignal Premise} {Julius Caesar was born on 12 or 13 July 100 BC Rome.  Julius Caesar died on 15 March 44 BC (aged 55) Rome.  The resting place of Julius Caesar is Temple of Caesar, Rome.  The spouse(s) of Julius Caesar are Cornelia  (84-69 BC; her death), Pompeia  (67-61 BC; divorced), Calpurnia  (59-44 BC; his death).}

\noindent \exampleParagraph{Orignal Premise + KG explicit}{ Julius Caesar died on 15 March 44 BC (aged 55) Rome.  \textbf{The resting place of Julius Caesar is Temple of Caesar, Rome. } Julius Caesar was born on 12 or 13 July 100 BC Rome.  The spouse(s) of Julius Caesar are Cornelia  (84-69 BC; her death), Pompeia  (67-61 BC; divorced), Calpurnia  (59-44 BC; his death). {\color{blue}KEY: Died is defined as pass from physical life and lose all bodily attributes and functions necessary to sustain life .} \textcolor{mdgreen} {KEY: Resting place is defined as a cemetery or graveyard is a place where the remains of dead people are buried or otherwise interred .} {\color{blue}KEY: Born is defined as british nuclear physicist (born in germany) honored for his contributions to quantum mechanics (1882-1970) . KEY: Spouse is defined as a spouse is a significant other in a marriage, civil union, or common-law marriage . }}

\noindent \exampleParagraph{Hypothesis}{ Julius Caesar was buried in Rome.}

\paragraph{Result and Explanation} 

\begin{table}[!htbp]
\small
    \centering
    \begin{tabular}{c|c} 
    \toprule
     Model & Label \\ \midrule
     Human Label (Gold) & Entailed  \\
     Original Premise & Neutral \\
     + KG explicit & Entailed \\
    \bottomrule
    \end{tabular}
    \caption{Prediction after KG explicit addition. Here, + represents the change with respect to the previous row.}
    \label{tab:exampleKGExplicit}
\end{table}

In this example from \alphaTwo, the model without explicit knowledge predicts neutral for the hypothesis as it is not able to infer that \textbf{resting place} is where people are \textbf{buried}, so it predicts neutral as it implicitly lack buried key understanding. On explicit KG addition (highlighted as {\color{blue} blue}+ \textcolor{mdgreen}{ green}), we add the definition of resting place to be the place where remains of the dead are buried (highlighted as \textcolor{mdgreen}{green}). Now the model uses this extra information (highlighted as \textcolor{mdgreen}{green}) plus the original key related to death (highlighted in \textbf{bold}) to correctly infer that the statement Caesar is buried in Rome is entailed.

\end{document}